# Dynamic Bayesian Networks to simulate occupant behaviours in office buildings related to indoor air quality

Khadija Tijani[1,2,3], Dung Quoc Ngo[4], Stephane Ploix[1], Benjamin Haas[2], Julie Dugdale[3]

[1]Univ.Grenoble Alpes, G-SCOP, Grenoble, France

[2]Centre Scientifique et Technique du Bâtiment , Champs sur Marne, France

[3]Univ.Grenoble Alpes, LIG, Grenoble, France & Univ. Agder, Grimstad, Norway

[4]Post and Telecommunication Institut of Technology, Hanoi, Vient Nam

## ABSTRACT

This paper proposes a new general approach based on Bayesian networks to model the human behaviour. This approach represents human behaviour with probabilistic cause-effect relations based on knowledge, but also with conditional probabilities coming either from knowledge or deduced from observations. This approach has been applied to the co-simulation of the CO2 concentration in an office coupled with human behaviour.

## INTRODUCTION

Most of the world standards for buildings take occupants into account by representative values and deterministic scenarios: number of occupants, predefined schedules, and responses to an exceeded threshold. Nevertheless, in low consumption buildings, the impact of occupant behaviour becomes a determining consumption factor because of its influence on energy consumption and indoor climate conditions through interactions with the building physics. For this reason, it is important, when estimating the energy consumption, to take into account occupant behaviour. For instance, if air quality is poor, occupants may decide to open windows or doors, depending on the type of windows but also on other occupants' activities or wishes.

A high performance building should be efficient for a large diversity of usages. Therefore, a need arises in building simulation: simulating the building with its appliances but also with reactive occupants i.e. with occupant behaviours sensitive to indoor conditions such temperature or air quality. Numerical models usually take into account occupant behaviour based on presence and on different profiles for opening windows. However, this may be insufficient since some simulation software handles the occupant model and the windows model separately, potentially leading to a situation where the occupant is absent, but the window is opened. In addition, profiles rely on data collected from specific observations. It makes it difficult to adapt them to other buildings or homes. To tackle these issues, this paper proposes a design methodology based on a knowledge model of occupant behaviour similar to building physics. It makes it possible to design occupants' behaviour using an a priori knowledge of future occupants and possibly using some observations to tune some model parameters. This model is then used for co-simulating physical and human aspects. The approach proposed in this paper relies on Bayesian Networks (BN) and has with advantage to represent human behaviour by probabilistic cause-effect relations based on knowledge, and also conditional probabilities coming either from knowledge or analysed observations.

This framework is applied to the co-simulation of an office combining a behavioural model representing occupancy and actions on door and windows, and a $CO_2$ physical models. The model of the human behaviour is tuned according to experimental data.

## STATE OF THE ART

In the scientific community of building physics, there is a growing interest for occupant behaviour because of its importance for energy waste reduction in buildings (Andersen et al, 2009). The occupant behaviour may be studied from building physics to human biology, through sociology and psychology in order to model and assess thermal, visual, comfort, and indoor air quality, etc. This paper focuses specifically on indoor air quality.

In the literature, three kinds of approaches can be found: deterministic approaches based on predefined scenarios and behavioural rules; statistical approaches that rely on factual observations such as surveys; and social modelling approach modelling cognitive and deliberative behaviours. In a deterministic approach, behavioural rules such as "if the temperature exceeds 28°C then someone opens the window" may be designed. However some researchers believe that human behaviour is better represented by stochastic models (Haldi et al, 2009). (Nicol et al, 2004) suggests that occupant behaviour corresponds to stochastic processes and not to deterministic ones: "there is not a precise temperature at which everyone opens his window, but the higher the temperature is, the higher the probability of opening the window is". (Yun et al, 2007) studies the probability of opening and closing window as a function of the indoor and the outdoor temperatures. The authors established a probability distribution of actions as a function of the indoor temperature. In 2009, inspired by previous works, (Haldi et al, 2009) proposed a hybrid stochastic model for window opening based on three modelling approaches: logistic probability distributions, Markov chains and continuous-time random processes.

More recently, a new approach to social modelling has been proposed that is based on agents. Here computational agents, each having their own characteristics and behaviours, are used to model the occupants. (Kashif et al, 2014) uses this approach to describe occupants' behaviour from an energy consumption point of view. The BRAHMS (Business Redesign Agent-Based Holistic Modelling System) platform has been used to model each agent's cognitive behaviour. BRAHMS is based on activity theory and is able to model an occupant's (agent's) beliefs, desires and intentions (BDI). BDI allows the representation of occupant's cognitive, reactive and deliberative behaviours. (Bonte, 2014) uses neural networks for the agents to learn their behaviour from recorded data to ensure their comfort. After a learning phase, agents know the actions that increase their comfort in different environmental conditions. In a recent study, (Langevin et al, 2014) models various occupancy profiles by calibration to predict the use of fans, heating and windows.

The social modelling approach is able to capture the same level of complexity as Markov chain processes. Combined with field studies, the agent-based approach proposes cognitive and deliberative schemas that go beyond statistical approaches. However, the complexity of social modelling makes it difficult to be used by designers (Tijani et al, 2014): a more focused modelling approach, described in the following section, would be more helpful.

## DYNAMIC BAYESIAN NETWORKS AS A GENERAL TOOL FOR HUMAN BEHAVIOUR MODELLING

A Bayesian Network (BN) is a probabilistic causal model represented by a directed acyclic graph. Intuitively, it is both a knowledge model and an inference engine using conditional probabilities and evidence to deduce resulting probabilities. Causal relationships between variables are represented by the graph. In the latter, the relationships linking causes to effects between variables are assigned probabilities. The observation of a variable does not automatically invoke the related effects but changes the probability of observing them. Because of the causal representation, a BN is a human friendly way to represent complex behaviours that can be summarized by conditional probabilities. The Static BN (SBN) only concerns a single slice of time. It is not relevant for analyzing an evolving system that changes over the time. Fortunately, a Dynamic BN (DBN) overcomes this problem by describing how variables influence each other over the time based on the model derived from past data. A DBN represents the relations between variables at different time steps, some of these variables can be not directly observed.

Compared to a BN, Hidden Markov Models (HMM) rely on nodes that correspond to states. Nevertheless, the possible number of states for human behaviour is numerous. For instance, in the studied case, 16 variables had been invoked to build the structure of the BN, each of these states has at least 2 states, that means $2^{16}$ combinations at least, with such a number, HMM cannot be used practically. BN is more general that HMM because each node stands for a variable (that could possibly be a state) corresponding to any fact, feelings, belief, desire or intention i.e. the basic concepts in social modelling. From a more technical point of view, (Oliver et al, 2005) states that the accuracy of inferences done by DBNs is less sensitive than HMM using the same amount of observations. Contrary to HMM, BNs offer an easy way to combine prior expert knowledge with conditional probabilities learnt from observations.

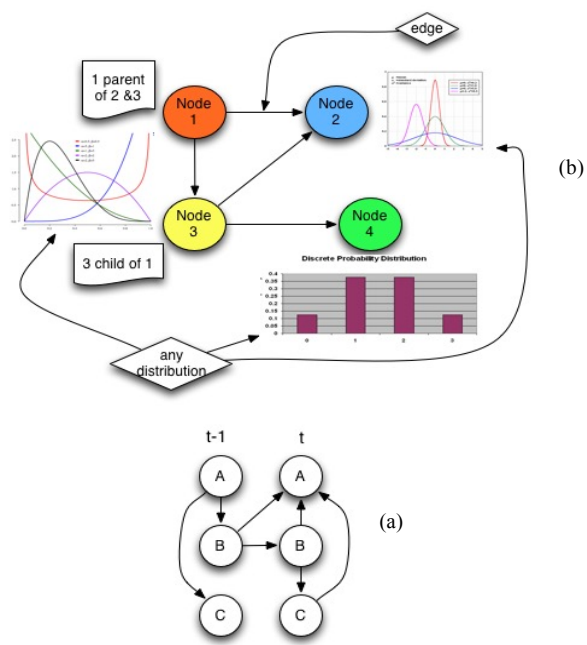

Fig.1 Elements of a Bayesian Network: (a) Dynamic BN (b) Different probability distributions

This point enhances significantly the precision in user behaviour modelling because merely learning a statistic model from data is generally not rich enough for user activities recognition.

To build a BN, a causal graph has to be defined as well as conditional probability tables for each node that can be tuned from knowledge or from observations. There are two approaches to design the structure of the causal graph. The first approach consists in learning the causality directly from data, if they exist, but sometimes, resulting causalities are not consistent with reality because data are too poor. The second approach is based on the design of a causal graph using expert knowledge about occupant cognition in different contexts and activities. The second approach is used in this work.

In BN modelling, any type of probability distribution function (PDF), such as uniform, normal, Poisson PDF (figure 1(b)), can be obtained with a directed edge between 2 nodes where the value domain of the input variable has been discretized. However, nonlinear functions of variables cannot be represented. It is therefore proposed to combine a DBN with nonlinear functions to process input evidence, for instance, the PMV model of the Fanger model, to process output inferences.

## CO-SIMULATION MECHANISM

This section focuses on the co-simulation mechanism. According to Bonte et al (2014), occupants act as controller of comfort, forming a loop summarized in figure 2. Therefore, the simulation is designed as a loop where the physical part plays, in this work, the role of an orchestrator, even if the occupant part or a external component could also play this role. Here, the occupant part contains a step methods, which is a method that manage the question/response between the physical part and occupant one. Because of the time step, actions and perceptions are averaged. For instance, instead of considering the natural 2 states 'opened' or 'closed' for a door, different states such as 'always closed', 'mostly closed', 'mostly opened' and 'always opened' must be preferred. For example, if the time step is one hour, sometimes the door is not closed for the whole hour, it may be closed just 30% of the time step and another door state like 'mostly opened' can be introduced.

It has been shown in the previous section that BNs are general tools for modelling human behaviour. Nodes in a graph could represent any element of a cognitive and deliberative model such those proposed in (Kashif et al, 2014), including perceptions, feelings, beliefs, desires depending on cognition and deliberations, intentions and actions. The occupant behaviour simulator is composed of a DBN editor, that uses the 'libpgm' library adapted to Python 3, with a library of existing DBNs representing elements of occupant behavioural models. It is thus possible to build the behaviour of occupants just as a home setting is built, taking into account the available knowledge about occupants and some assumptions, or reusing existing models. A composition tool makes it possible to combine model elements. A scenario editor can generate evidences for different times and dates. Databases of scenarios and results, as well as a result viewer, are available.

The proposed approach couples occupant behaviour with an air quality energy performance calculation tools, which rely on dynamic energy calculation rather than on pure thermal calculation. Indeed, without energy calculations that take into account HVAC systems, the coupling with the occupant is less informative. The work is based on an existing tool from CSTB named COMETh (An introduction to development of the French Energy Regulation indicators and their calculation methods) that already includes models for a wide range of HVAC systems (mostly those modeled in the EPBD CEN set of standards). Therefore, coupling occupant behaviour with physical aspects is implemented by overriding the control modules by a consistent occupant module.

The actions of the occupants on the air quality must be narrowed down to some very specific critical points. Because the occupant models are coupled with physical models, the occupants' actions that must be modelled are constrained by the physical computations. These values are: a Boolean indicating whether the space is occupied or not, metabolic gains per thermal zone, activated appliances with corresponding gains per thermal zone, metabolic humidity gains per thermal zone, set point temperature, heating and cooling period if managed by the occupant, ratio of window opening and ratio of door opening.

An occupant model must provide all of these values. As already stated, because actions and perceptions are averaged, transient states must be carefully designed. In addition, the physical simulation must provide the quantities that allow the occupant to make decisions and take actions. A minimal set of data, that can enrich the simulation, is the operative mean temperature, mean humidity and $CO_2$ concentration.

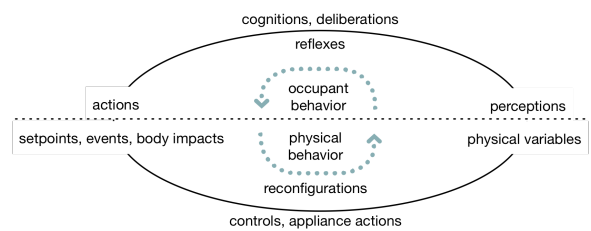

Fig.2 Elements of the model

# MODELLING OCCUPANTS IN AN OFFICE SETTING

The chosen office is non-mechanically ventilated and has several sensors to provide data. The sensors capture indoor temperature, motion, $CO_2$, door and window positions and humidity. Video recording equipment is also used to track occupant movements and actions. The office can be occupied by up to 4 persons. The possible occupants are: a Professor, a permanent PhD student, an intermittent PhD student, a guest and a visitor. Each of these occupants has his own schedule and behaviour. The professor is not regularly present in the office, but when he is, he works on his computer and usually remains seated. He often closes the door for privacy and welcomes visitors in his office. The permanent PhD student is always in the office and she often doesn't care about the door state unless to keep the office warm. The intermittent PhD student is at the office one week out of two, her desk is far from the door so she didn't care about it unless there is a noise in the corridor. The guest (for example an internship) has a desk near the door, so for his privacy, he prefers to close the door. The visitor is not an occupant of the office, but he comes to it to have meeting with one of the occupant, he mostly let the door as he found it. Only the professor and the permanent PhD student manage the windows, it's because of the proximity and also because of their status: they are permanents, they act intuitively and normally in the office.

This information is taken from the analysis of the video data. The DBN structure is shown in figure 3. In this structure, the target nodes are "Door" and "Window". The first one is influenced by six variables, which were identified through video data, each node has its own conditional probabilities. The value domains of variables are represented in figure 3. This structure is dynamic because it uses the "past" state, which is shown in figure 3 by an arrow from and to the same variables: door and window. The nodes "Calendars" are based on real occupant calendars, it is taken from their Google ones, so it can helps to build the structure in the way shown on figure (3). The level of $CO_2$ can be low, medium or high. The Professor's activities depend on the presence of the other occupants, he can be out, he can work alone on his computer, he can be in meeting with one or several occupants, he can be in virtual meeting on his computer. These activities influence the door movement not the window, based on the video observation. The occupants of the office are divided into two groups: one, which the opening of the door troubled, and the other one not troubled. The weather influences both, the door and the window. To make decision about the openings, the occupants deliberate between them. The conditional probabilities of the appropriate nodes are filled so this deliberation can be illustrated in the BN model. A tool (figure 4) has been developed in Python 3 to manage the creation of the structure and the input of conditional probabilities. It relies to an adaptation of libpgm module. First, the nodes are created: name, causes, possible values and conditional probabilities. The tool can also calculate the Bayesian inferences at each node for a given time and then propagate the results as evidence into the next time BN. All conditional probabilities, in this work, are empirically determined, based on the video observation and office occupants questioning. It can also be learnt using virtual estimator but it is not going to be developed in this paper.

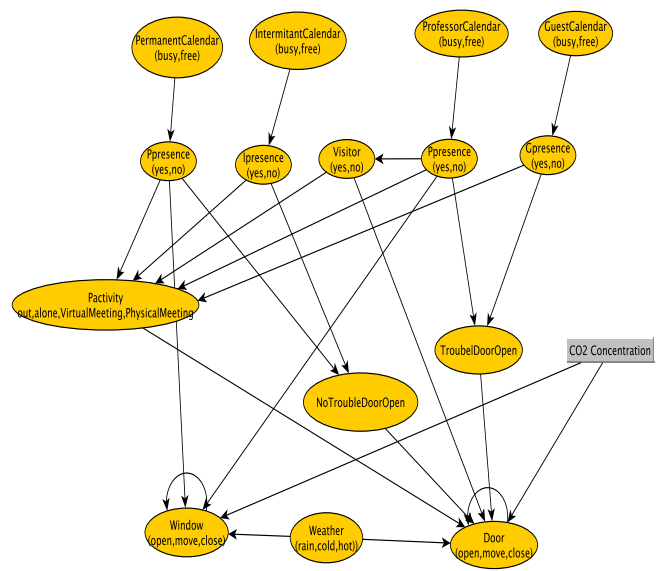

Figure 3 - The dynamic Bayesian network (DBN) structure

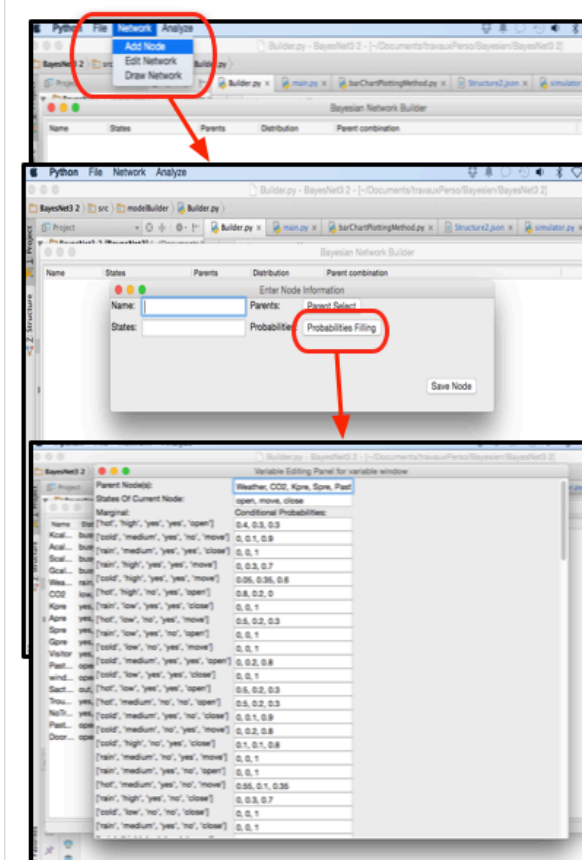

Figure 4  DBN structure creation tool

## PHYSICAL MODEL OF THE OFFICE SETTING

The aim of this model is to predict the rate of $CO_2$ concentrations in the office depending on the door opening. To design the model, data have been collected from sensors as well as physical models of $CO_2$ generation (Aglan et al, 2003) and flow calculations (Mounajid et al, 1989). The equation (1) is based on mass balance considerations in a zone with a flow of air renewal.

$$C_{k+1} = \left(\frac{Q_i}{Q_o}C_i + \frac{S}{Q_o}\right) - \left(\left(\frac{Q_i}{Q_o}C_i + \frac{S}{Q_o}\right) - \ldots C_k\right) e^{-\frac{Q_o}{V}(t_{k+1}-t_k)} \quad (1)$$

where $C(t)$ is the $CO_2$ concentration inside the building at time $t$, $C_i$ is the $CO_2$ initial concentration, $V$ is the office volume, $S$ is the generation rate of $CO_2$ from occupants exhaled air, which is a function of the volumetric rate of $CO_2$ (ASHRAE 1985). $Q_i$ and $Q_o$ are, respectively, the volumetric flow rate of renewed air. It is calculated by equation (2), which calculates the airflow passing through a large opening between the corridor and the office.

$$Q_{i,o}(z_1, z_2) = \frac{2}{3\rho}\epsilon CL\sqrt{2\rho|\Delta\rho|g}\left(|HN - z_1|^{3/2} - |HN - z_2|^{3/2}\right) \quad (2)$$

where $Q_{i,o}$ is the incoming or outgoing air flow, $\epsilon = \Delta T/|\Delta T|$, $\rho$ density of air (depending on the corridor and office air densities), $g$ gravity, $HN$ neutral plan (when the pressure difference between two areas is 0), $z_{1,2}$ are the heights of the opening that can be $HN$, low or high height depending on the temperatures.

The carbon dioxide is discretized into 3 values:
   -low : < 1000
   -medium : <1700 and > 1000
   -high : >1700

The occupant first reacts at the medium level; he starts breathing heavily and some times having headache. When the $CO_2$ rate reaches, or exceeds, the high level, the occupant may have more containment problems, like bad smells, migraines…

## RESULTS

To illustrate the proposed approach, the occupant calendars are represented in figure 5. It has been chosen that the weather is "hot" with some drops of rain between noon and 2pm (figure 7(a)) and without rain (figure 7(b)).

This working day has been generated by co-simulating 100 times the Bayesian network together with the air quality model yielding the results in figure 5.

As said before, this office has 4 occupants and 1 visitor. It means that the rate of the carbon dioxide (figure (5)) is changing very often and it may take a high level, and also means that the door is all the time used. Figures 6(a&b) show that the door is more in the state "Move". In figure 6(b), the door at 1pm is more used than at 12am. It shows that some occupants come back to the office after the lunch break. When it is raining, the window is totally closed because the rain uses to come into the office; so all the occupants close the window when it is raining.  The time slot 11am in both simulations shows that if one occupant is free, the door is more used than the other time slots when everybody is busy. Comparing also the time slots 17h and 12h, which correspond both to breaks (lunch and coffee break), it can be seen that occupants move inside the office (coffee break) or outside (lunch) because of the door uses. The time slots between 15h and 17h in figure (5) show an increase in $CO_2$ concentration, and it matches with the simulation's figures, all the occupants are busy, the door and window are not very used and may be a visitor is in the office.

These results clearly show that the proposed DBN can handle, on the one hand, the differences between two same inputs (here free) but with different activities, and on the other hand, the differences between acting on a door and on a window, and at the end the results show that there is deliberation between the occupants.

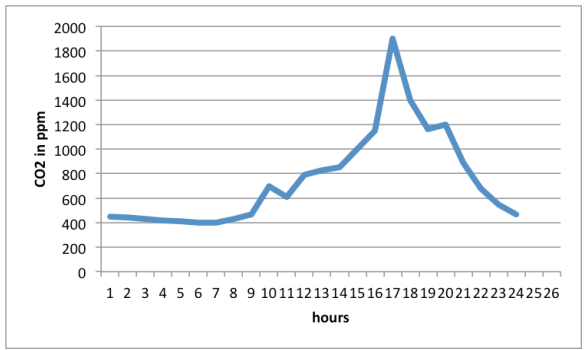

Figure 5 Carbon dioxide rate in one day

|  | 7h | 8h | 9h | 10h | 11h | 12h | 13h | 14h | 15h | 16h | 17h | 18h | 19h | 20h |
|---|---|---|---|---|---|---|---|---|---|---|---|---|---|---|
| PrCalendar | free | free | busy | busy | free | free | free | busy | busy | busy | free | free | free | free |
| Pcalendar | free | busy | busy | busy | busy | free | free | busy | busy | busy | free | free | busy | busy |
| ICalendar | free | free | busy | busy | busy | free | free | busy | busy | busy | free | busy | busy | free |
| GCalendar | free | free | busy | busy | busy | free | free | busy | busy | busy | busy | free | free | free |

Figure 6 - 24h calendars of the 4 occupants

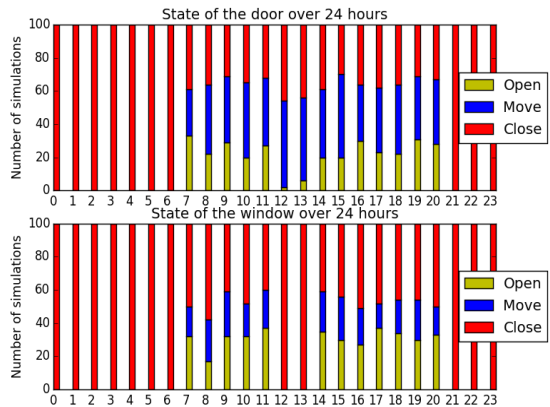

Figure 7(a) - 100 co-simulations of one working day for the 4 occupants in case of rain at 12h

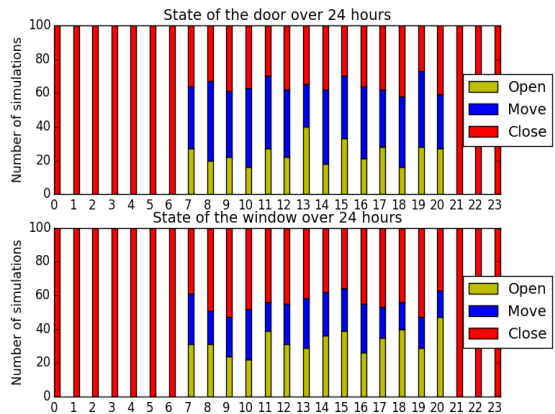

Figure 7(b) - 100 co-simulations of one working day for the 4 occupants in case of no rain at 12h

## ANALYSIS

The proposed approach combines expert knowledge to build up the structure of dynamic Bayesian networks and statistics coming from observation. Expert knowledge yields cause-effect relations: it is based on observations and by questioning occupants to understand the underlying causes behind actions. The conditional probabilities can be set up by expert knowledge, analogy with similar situation or by using sensors and virtual sensors to estimate values at the no measurable nodes of the dynamic Bayesian networks. The proposed approach can use statistics as with alternative human behavior modeling approaches, but in addition, expert knowledge can be introduced to take into account specific knowledge about occupants and future occupants as illustrated in figure 3. The Bayesian network formalism leads to the discretization of value domains of nodes and therefore to the set up of numerous conditional probabilities, more than 1000 probabilities have been filed in this work case, but as advantage, it offers a common framework to represent any kind of distribution functions.

Comparing to occupant profiles, the resulting human behavior model is reactive. Indeed, the behavior depicted in figure 3 depends on the weather but also on indoor CO2 concentration i.e. on the feeling of confinement.

The proposed approach takes advantage of the cognitive and social modeling concepts developed in (Kashif et al, 2014). Indeed, homeostasis, beliefs, desires and intentions can be taken into account, including for deliberative and social behavior among a group as show in figure 3, thanks to specific nodes. But it is also possible to not go that deep into the representation of reasons behind actions using overall probabilities. It is up to the designer to determine how far to go into details.

## CONCLUSION

A new approach based on a hybrid DBN has been proposed to model human behaviour. Causal representation eases the modelling process, as it is a natural way of thinking about human behaviour. Furthermore, a probabilistic modelling approach provides a relevant level for representing human behaviour, which becomes very complex at a finer level. Human behaviour can then be designed step-by-step, just like the physical part of a building system, taking into account the knowledge about future occupants and possible assumptions about their behaviours. This approach has the advantage of being intuitive, like cause-effect human reasoning. It also takes into account both the knowledge model and the probability distribution functions, which is missing in other approaches.

This approach has been illustrated in the co-simulation of $CO_2$ concentration with human behaviour in an office. It will be extended by performing co-simulations with dynamic energy simulation software.